# A quantitative investigation for deployment of mobile collaborative robots in high-value manufacturing

**Amine Hifi[1*], W. Jackson[1], C. Loukas[1], M. Shields[1], A. Poole[1], E. Mohseni[1], C. N. MacLeod[1], G. Dobie[1], S. G. Pierce[1], T. O'Hare[2], G. Munro[3], J. O'Brian-O'Reilly[3], R.W. K. Vithanage[1]**

[1] SEARCH: Sensor Enabled Automation, Robotics & Control Hub, Centre for Ultrasonic Engineering (CUE), Department of Electronic & Electrical Engineering, University of Strathclyde, Royal College Building, 204 George Street, Glasgow G1 1XW.
[2] Spirit AeroSystems Belfast, Airport Road, Belfast, Co. Down, Northern Ireland, BT3 9DZ
[3] Spirit AeroSystems, Aerospace Innovation Centre, Glasgow Prestwick Airport, Monkton, KA9 2RW

**\* Correspondence:**
Amine Hifi
amine.hifi@strath.ac.uk

## Abstract

Component inspection is often the bottleneck in high-value manufacturing, driving industries like aerospace toward automated inspection technologies. Current systems often employ fixed arm robots, but they lack the flexibility in adapting to new components or orientations Advanced mobile robotic platforms with updated sensor technologies and algorithms have improved localization and path planning capabilities, making them ideal for bringing inspection processes directly to parts. However, mobile platforms introduce challenges in localization and maneuverability, leading to potential errors. Their positional uncertainty is higher than fixed systems due to the lack of a fixed calibrated location, posing challenges for position-sensitive inspection sensors. Therefore, it's essential to assess the positional accuracy and repeatability of mobile manipulator platforms. The KUKA KMR iiwa was chosen for its collaborative features, robust build, and scalability within the KUKA product range.

The accuracy and repeatability of the mobile platform were evaluated through a series of tests to evaluate the performance of its integrated feature mapping, the effect of various speeds on positional accuracy, and the efficiency of the omnidirectional wheels for a range of translation orientations. Experimental evaluation revealed that enabling feature mapping substantially improves the KUKA KMR iiwa's performance, with accuracy gains and error reductions exceeding 90%. Repeatability errors were under 7 mm with mapping activated and around 2.5 mm in practical scenarios, demonstrating that mobile manipulators, incorporating both the manipulator and platform, can fulfil the precise requirements of industries with high precision needs. Providing a highly diverse alternative to traditional fixed-base industrial manipulators.

## 1    Introduction

Mobile robotic platforms, particularly for warehouse management, are frequently employed for moving goods between stations. Mobile manipulators combine a mobile platform, with an onboard manipulator. Currently, mobile manipulators have not been extensively used in high value manufacturing applications, operating autonomously at full capacity. Several factors have impacted the adoption of mobile manipulators, including the costs associated with their implementation.

Additionally, the shortage of qualified personnel to operate these systems and the necessity for updated safety protocols to handle the systems correctly are also significant factors (1).

While fixed arm robots are widely used for automation of cumbersome and repetitive procedures such as welding, metrology, and machining in high volume production lines, they are relatively inflexible when it comes to adapting to new components, different component orientations, or new operations (2). As a result, the environments in which they operate tend to be designed to accommodate the robot's limitations. This is why some businesses opt for manual labour methods instead, especially when dealing with high-mix and low-volume production components (3). Additionally, reach limitations require installations of additional external axis to introduce more DoF (Degrees of Freedom) and machining of appropriate gantries or fixtures to handle the components. For instance, aerospace components are often large (>3 m) and necessitate customized systems for manufacturing processes (4) where such systems lack the adaptability and flexibility introducing delays in manufacturing and impacts production throughput.

Therefore, new methods and processes must be designed to be able to adapt to all aspects of the manufacturing operations. The introduction of collaborative mobile robotic platforms is one possible solution. Utilizing modern sensor technologies and algorithms, mobile robotic platforms have shown increased ability and accuracy in localisation and path planning making them well-suited for bringing the inspection process to the part. Mobile systems introduce greater flexibility and independence from infrastructures, such as gantry-based systems, decreasing the amount of capital equipment required to accommodate components.

The primary objective of this study is to evaluate the suitability of mobile manipulators for high value manufacturing processes with special focus on Non-Destructive Testing (NDT). NDT is a non-intrusive method used to evaluate the properties of materials, components, or systems. NDT techniques include radiographic testing, which uses X-rays or gamma rays to penetrate materials and detect defects. Ultrasonic testing detects defects using sound waves, magnetic particle testing detects surface and near-surface flaws in ferromagnetic materials, and visual inspection visually assesses materials for defects. By using NDT, defects and irregularities can be detected early on, helping to prevent accidents and costly failures. Due to the need for precise and reliable results in these procedures, an in-depth analysis of mobile manipulator capabilities is warranted, accompanied by the establishment of attributes that define their performance.

By implementing a mobile manipulator solution, the manufacturing process can be streamlined significantly (5). A mobile solution that can work with a range of sensors and tools, with the capability to travel to different sections of the production line on a factory floor, means there is no longer any need to have several divided areas of a factory, when the process can come to the part. This can therefore reduce a factory footprint significantly, introducing reductions in costs for overheads and equipment. Mobile manipulators have the ability to contribute throughout the entire lifecycle of manufactured goods, from production to disposal.

Industry 4.0 is synonymous with smart manufacturing and has brought about the necessity for the digital revolution in this sector. This entails the use of artificial intelligence, big data analysis, and other advanced technologies to create interconnected systems that operate more efficiently, autonomously, and effectively (6). Growing demands and skilled labour shortages have also contributed towards the growing trends of automation across various industries (7,8). Manufacturing sectors, in particular, having the highest potential for automation (9). While automation improves job efficiency and has economic benefits, not all tasks can be automated, creating a need for hybrid work environments combining manual labour with automation (9). Mobile manipulators are seen as a flexible solution to bridge this gap, as they can navigate and adapt to different environments and tasks effectively.

To understand the capabilities of mobile manipulator platforms and hence to investigate their performance for different end-user applications, it is necessary to consider a specific operation scenario. The field of NDT demands a high level of accuracy and adheres to strict inspection standards (10). Ultrasonic Testing, UT, is a form of NDT and it is crucial to conduct UT precisely as even minor changes in angles can lead to missed defects. Correct surface contact and coupling is also crucial for UT. Improper coupling can negatively impact signal strength, sensitivity, and resolution, leading to inaccurate measurements and potentially missing defects in the test material. For critical applications such as detecting defects in welds or assessing material thickness in aerospace or nuclear industries, the required accuracy is often within the range of fractions of a millimetre or a percentage of material thickness. This study focuses on evaluating the accuracy and repeatability of a mobile robotic manipulator for its applicability in the NDT sector.

At present, the International Organization for Standardization (ISO) standard ISO 24647 sets out the criteria for robotic ultrasonic testing, but it does not include mobile manipulators (11). ISO standard ISO 9283 establishes standardized testing methods for accuracy and repeatability in industrial manipulators (12). Yet, there is no existing standard for mobile manipulator platforms in terms of repeatability, accuracy procedures, or the specific needs for non-destructive testing applications.

Therefore, the aim of this work is to complete a set of tests that will help determine whether the mobile manipulators can achieve the necessary precision required for automated inspections. There are several popular mobile platforms that were considered such as the RB-KAIROS+ and ER-FLEX(13,14). These platforms have a range of features such as built in vision systems and integrated Universal Robot arms onboard (15). The KUKA KMR iiwa (16) was selected for these experiments, due to its robust build, integrated laser sensors, omnidirectional Mecanum wheels, and the scalability of KUKA with its variety of robot systems available - shown in Figure 1. The scalability within the KUKA product range presents the opportunity for the lab-environment developments to be transferred to larger industrial robots.

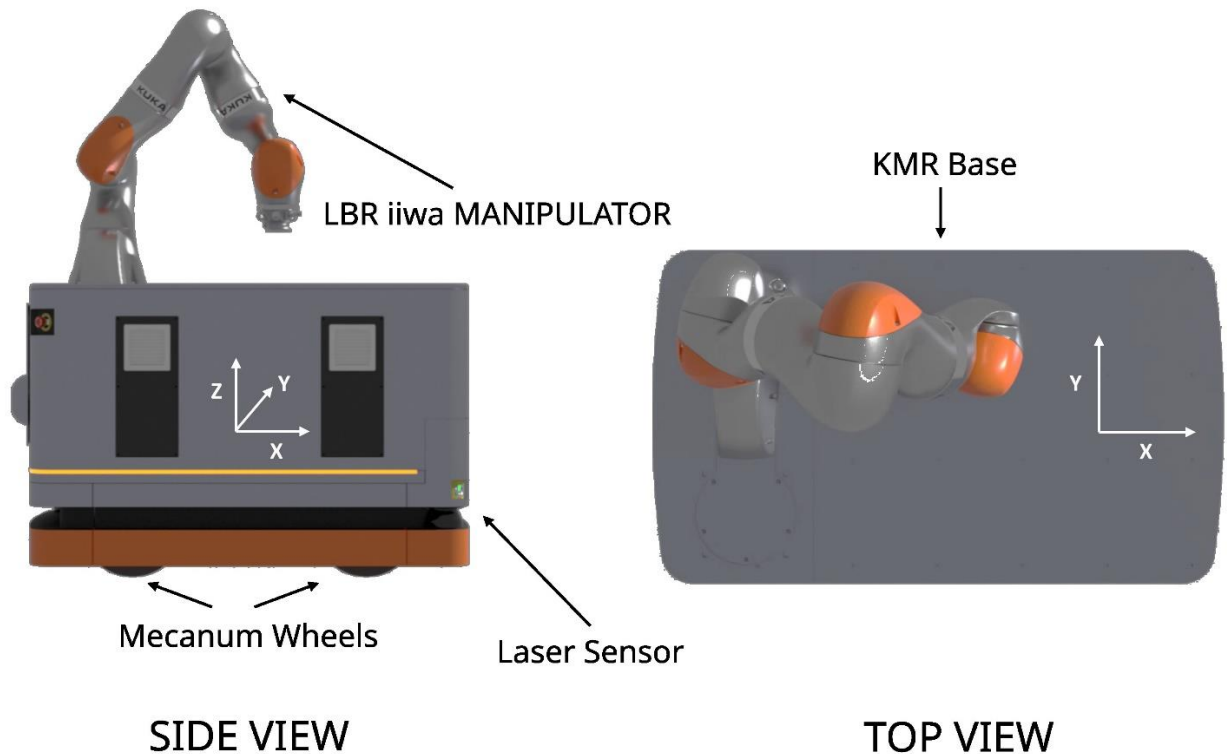

Figure 1. KUKA KMR iiwa mobile manipulator (The coordinate frame refers to the KUKA KMR base frame)

Depending on the purpose and context, NDT can be approached in various ways. In the oil and gas industry, asset monitoring solutions often utilize crawlers and drones to improve flexibility and accelerate the inspection process (17,18). Crawlers with ferromagnetic wheels are preferred in situations where the materials are ferrous. In the case of composite materials, often seen in aerospace, automotive, and energy sectors, the current methods are to inspect the parts in a fixed gantry, using a UT water jet or UT immersion scanning with robots (19). These systems can achieve very precise and repeatable measurements if set up correctly. Robotic manipulators are currently sufficiently accurate for many high value manufacturing applications, as some manufacturers claim repeatability values of ± 0.1 mm (20).

ISO defines a mobile robot as, "*a "robot able to travel under its own control". A mobile robot can be a mobile platform with or without manipulators. In addition to autonomous operation, a mobile robot can have means to be remotely controlled*" (21)

Crawlers are a prevalent example of mobile robotics extensively utilized in NDT. One such system used a cylindrical shaped crawler, for the inspection of defects using a vision system and lasers, traversing through pipes (22). The system was able to move through 100 m of piping, with a diameter of 100 mm. One of the takeaways of their system was that the mobile robots should be able to travel significant distances to be able to be used in industry. It is crucial to acknowledge and recognise the accumulation of errors that typically occur with long travel distances, which stands as the most significant aspect requiring attention. Quantifying and understanding these are essential.

To travel across a factory floor, typically dynamic and unstructured, the mobile platform must have a more complex set of sensors in comparison to the crawler platforms. Integrated SICK S300 laser sensors (23) equipped with the KUKA KMR iiwa platform facilitates navigation through different environments, as these scanners serve the purpose for both mapping and ensuring safety.

The KMR platform has been used in many cases due to its collaborative nature. Making use of its mapping features and the manipulator onboard the robot, the KMR could open and traverse through doors (24). The simultaneous localization and mapping approach, SLAM, along with Adaptive Monte Carlo Localization (AMCL), allowed the KMR to create a map of its environment and accurately locate itself within it. The AMCL algorithm uses particles to represent the robot's possible configurations and adjusts their weights based on sensor data to converge towards the actual position. To improve accuracy, additional laser scans are taken in front of the door and the robot aligns itself with the door to transform its frame relative to the door's frame. This strategy of re-aligning after initial movement can be applied to high value manufacturing for accurate platform positioning.

The KMR was used in experiments to understand its collaborative robotic limitations (25). In tests, a bar was moved from one side of the laboratory to another, with the operator gripping one side and the robot on the other. The robot adjusted its movement based on haptic feedback from the operator, achieving accuracy measurements with error deviations of "0.01257% and 0.006%", along the x and y axis respectively. However, the setup and details of the experiment were not well-described. Overall, the study concluded that humans and robots can effectively work together, with the robot enhancing task efficiency and performance.

The accuracy of the inspection process relies on precise positioning and scanning to identify defects accurately. The performance of the mobile platforms can vary, depending on the environment and surface conditions. A vision system has been suggested as a solution to address positional uncertainty. It has been identified as the most effective approach for dealing with positional uncertainty (26). The concluding remarks found that the vision system was able to achieve compensation with a 0.19 mm offset outperforming the likes of fine positioning and lidar feedback as compensation methods for positional inaccuracies.

A dual arm mobile robot was investigated for its ability to navigate a shop floor (27). Utilising the SLAM algorithm solely left the robot out of position by 50-100 mm at times. By introducing AR Tags and a vision system, these errors were reduced to 10 mm or less. The concept of achieving a significant reduction in positional errors by more than 80% through the implementation of an integrated vision system has proven to yield substantial improvement for both existing and for possible future mobile systems.

The onboard manipulator, LBR iiwa, is capable of movements with a repeatability of <0.1 mm (28). Its usage has also been observed in the medical field, specifically in a knee arthroplasty procedure where the manipulator operated in compliance mode (29). Consequently, these mobile manipulators possess the ability to handle delicate tasks and execute accurate measurements.

Several researchers have investigated the validation of the positional accuracy for different platforms. One research team proposed the use of an L-shaped approach to validate the accuracy of their mobile platform (30). This was found to be too tailored to meet the specific expectations of the use case, without any translation to the KMR iiwa platform testing or quantification. Other groups have looked at testing their robots with circular paths and figures of eight (31,32). The results were determined by finding the displacement of the robot after these movements. This work aimed to adopt the suggestion of examining the overall displacement after the paths were completed. However, the paths tested by other researchers appeared unfeasible because of the size difference between the KMR and the robots used for these tests, and effective experimental space limitations (30),(31,32).

The present NDT environments utilizing robotic manipulators are not collaborative in their methods, and still lack the essential technologies to accomplish fully autonomous NDT inspections without human intervention.

Currently, there is no research that has presented the use of mobile manipulator platforms for NDT that possess complete autonomy to freely navigate and explore their surroundings. To get baseline values required for a suitable mobile manipulator, the specifications of an in house designed phased array ultrasound roller probe is being utilised (33). For this roller probe there is a fundamental requirement of a high degree of precision to achieve the expected results. From previous studies, experimental validation has deduced that using such sensors necessitates advanced control capabilities (33,34). For optimal results, it is necessary to have contact forces with accuracies of 0.5N, positional accuracies that are better than 0.5mm, and rotational accuracies that are better than 0.1 degree. Other commercially available phased array ultrasound roller probes perform similarly and are expected to require comparable levels of accuracy.

The unique aspect of this work is its approach to designing experiments that assess the precision of mobile robotic platforms, specifically in the context of evaluating the potential use of mobile manipulators in high accuracy NDT applications.

## 2    Experimental Design

Since there is no standard in place to establish and quantify the capabilities of mobile robots, this paper proposes a new novel strategy, inspired by the ISO 9283 (12), to examine various aspects of the mobile manipulator's capabilities. For the KMR that is utilised in this study, this includes:

a) Assessment of the mapping features,

b) Understanding the impact of different speeds on the positional accuracy of the KMR,

c) Test the operational capabilities of the Mecanum omnidirectional wheels

The specification for the testing was to ensure that the different aspects and features of the mobile manipulator platforms were tested.

The four mobile platform Mecanum wheels with three Degrees of Freedom (DoF) enables unrestricted motion and orientation of the platform in two-dimensional (2D) space. Strafing movements can occur with the use of Mecanum wheels, in which the robot will move in any direction without changing its orientation. During strafing the platform can complete translations in any direction, without doing a full rotation offering great flexibility in confined environments.

Therefore, given that the KMR can translate to a preset position on the floor either using strafing, or a combination of translation and rotation, one logical comparison point for this Mecanum omnidirectionally enabled platform, is to compare strafing capabilities, Figure 2 a), versus full rotations, Figure 2 b). The KMR is also capable of driving with solely the encoders fitted onboard, without relying on the laser scanners for laser mapping. For this reason, a plan was devised to test the motions of the KMR during the strafing and rotations using its encoders as compared against the motions with mapping features enabled.  For mobile platforms that lack holonomic capabilities, it is suggested that rotational tests will still yield adequate results concerning accuracy and repeatability capabilities.

The testing environment was a 3-meter by 3-meter floorspace, to accommodate the size of the KUKA KMR platform. The floor was an SR1 Floor constructed with epoxy resin, finished with Polyurethane Floor Paint. Within the test area the floor was measured at $0°$.

The final path design was constructed in the shape of a square with 1 $m^2$ in area to test the strafing and rotational capabilities. The path design is demonstrated in Figure 2 a), with target positions all spaced 1 m from each other. Moreover, the 4 m square test path provided a uniform distance between all target positions and functioned to ensure that the distance travelled accurately reflected the intended 1-meter target for each path. The rotational test followed the same structure as the strafing, shown in Figure 2 b).

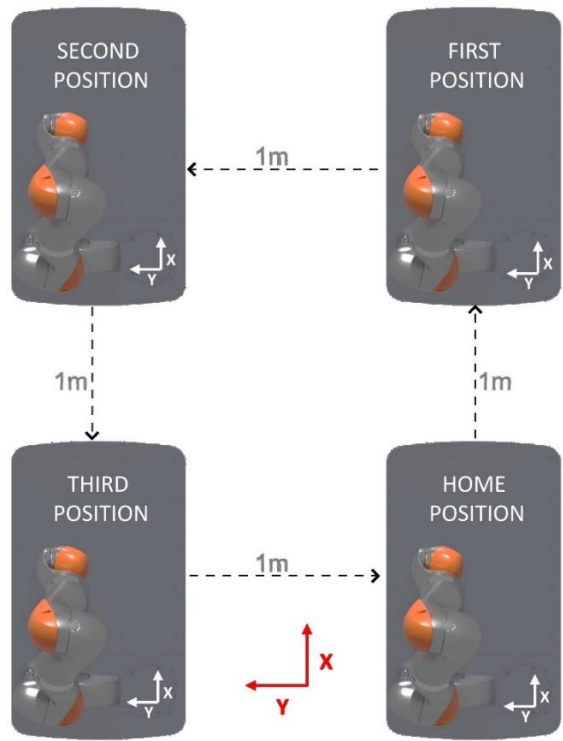

a)

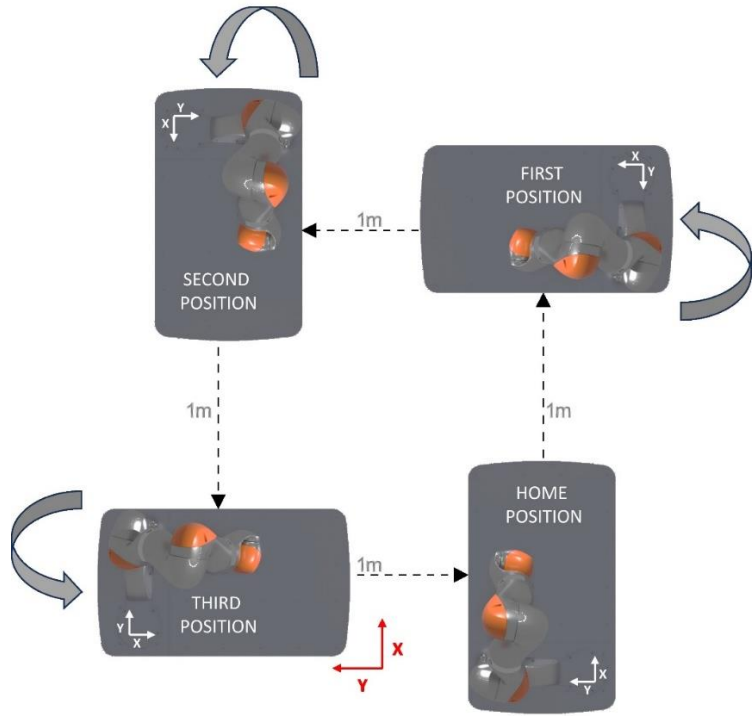

b)

Figure 2. Planned Path Positions. a) Strafing, b) Rotational (Global frame in red, KMR frame in white)

## 2.1    Position Tracking

A laser tracker was used to track the performance and movements of the KMR. The tracker works by emitting a laser beam onto a target reflective surface on the object being measured. The laser beam is then reflected back to the tracker, allowing it to calculate the position of the target in three-dimensional space. The Leica Absolute Laser Tracker AT901(35) was selected because of its 10 µm precision that can precisely determine the absolute position and angle of the KMR within a $160m^3$ volume. Its 3 DoF capability ensures accurate tracking at different heights and angles, while the laser reflector on the KMR enables the tracker to measure the position and orientation. If the line of sight of the Leica is broken, all tracking is terminated. In strafing tests, the platform is always facing the same direction, meaning there were no concerns about the laser signal being disrupted, as demonstrated in Figure 2 a). However, the manipulator arm onboard interrupts the laser signal when the platform rotates more than 90 degrees, which presents a problem for rotational experiments.

For this reason, a rotation stage was designed to be placed atop of the manipulator, ensuring that its position would always be uninterrupted. This rotation stage was designed to be capable of remaining focused on one point and rotating accordingly in $\pm 2\pi$ to always face a "true north". The Vicon T160 camera system (36) was used to determine its true north position and adjust the rotation stage orientation accordingly. The Vicon camera system, is a 3D motion capture system with 12 cameras mounted surrounding a volume. In this method, infrared cameras are used to track retroreflective markers. Through experimentation, it has been determined that the system has a calibrated absolute error of 0.51mm (37).

The KMR platform had these reflectors mounted on board, in several different positions to allow for the Vicon system to detect and track it. Once initialised and calibrated the KMR position in the volume was tracked and followed by the Vicon system. The Leica laser reflector ring has a tolerance $\pm 30^o$, therefore the rotation stage does not require the highest degree of accuracy. The laser reflector was mounted to the rotation stage using a machined aluminium mount. The holder is made with an inbuilt magnet to hold the reflector rigidly. This approach provides the ability to rotate the reflector and always be able to accurately measure the position of the KMR platform.

### 2.1.1 Leica validation

To validate the accuracy of the Leica laser tracker, the reflector was placed in a single position for an hour while the laser tracker recorded every 100 ms. 37651 entries were recorded equating to just over 62 minutes of run time, for validation. This was done to verify the variation in the positional measurements. The results of this initial validation gave further confidence that the Leica could measure to a very high precision degree. The standard deviation was measured at 4.2165 µm, 1.8931 µm, 5.2145 µm for X, Y and Z components respectively.

## 2.2    Controlled Features

The 2D laser scanners onboard the KMR platform allow the robot to localize and navigate using Simultaneous Localization and Mapping (SLAM) (38).

For laser mapping, there is a need to keep a constant consistent environment, any changes in the layout can impact the mobile manipulator's ability to localize. The SLAM utilisation for tracking and mapping in a dynamic environment has been explored previously (39), but its widespread adoption is still pending. This study only focuses on benchmarking the capabilities of the KMR platform, therefore no third-party SLAM algorithms were explored or incorporated into the experiment. For this reason, adding consistent surroundings that are fixed can help to find the capabilities of the onboard SLAM mapping of the KMR. In this situation, cardboard was used to surround the work area as it is a non-reflective material that provides a solid planar barrier framework. Three walls were installed, using non-reflective panels leaving the last side open to allow the KMR to accurately determine its position. This technique was used to avoid confining the KMR in a box with essentially the same wall features, which could confuse the localizing abilities.

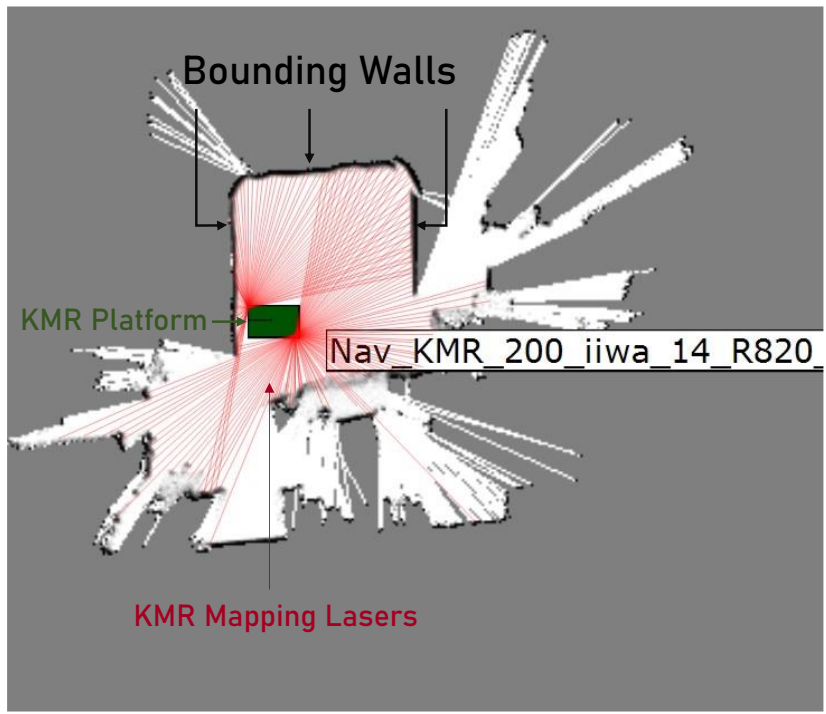

Figure 3. KMR lasers identifying position in SLAM map.

With these controlled features, the KMR would be able to identify its position. The implementation of the controlled features resulted in the identification of the three walls and the fourth side open as seen in Figure 3. The final experimental setup is presented in Figure 4.

The experimental design plan containing controlled features aligns with the ASTM standard F3244-21 for Navigation: Defined Area (40), in which individual tasks are repeated multiple times by the mobile platforms, within a designated floor space outlined by physical barriers.

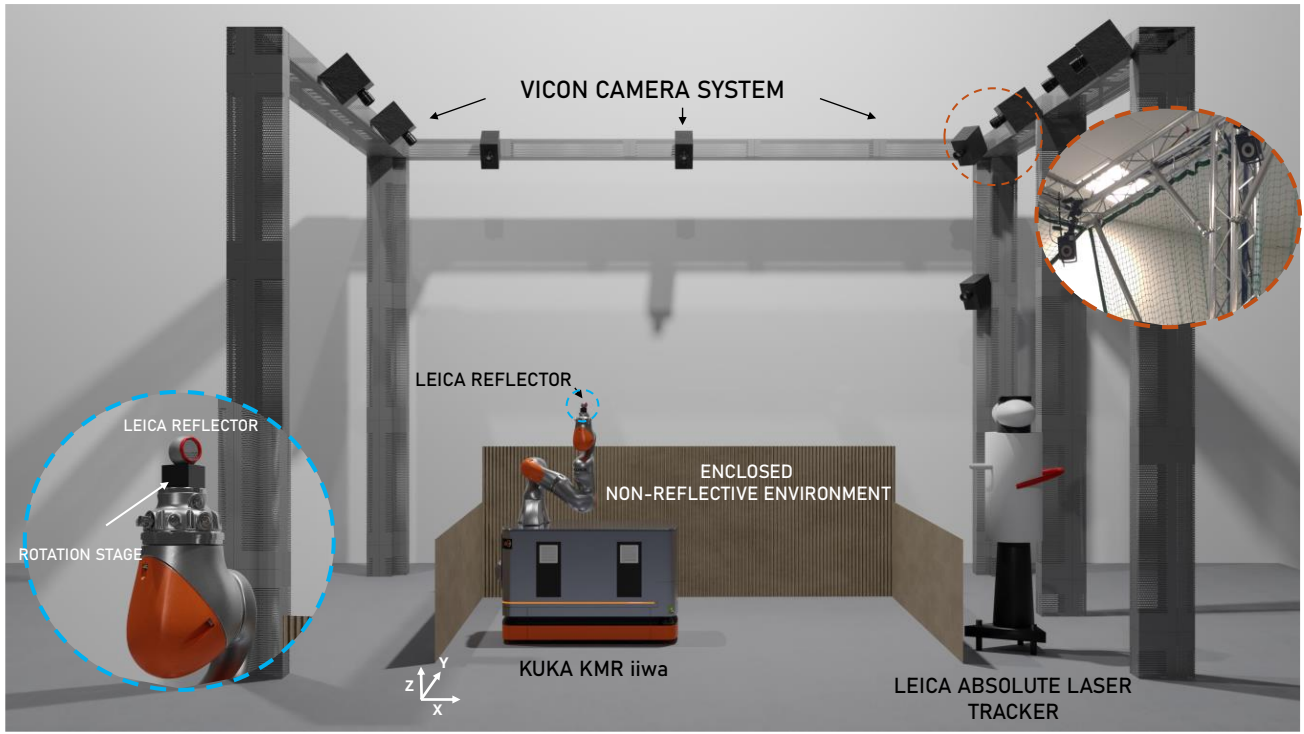

Figure 4. Experimental setup for positional tracking of the KUKA KMR iiwa utilizing Leica laser tracker. (Co-ordinate frame represents the global frame for measurements)

## 2.3   Teaching map and points

The KMR, with its lasers in mapping mode, is manually jogged in an enclosed space in order to teach the mobile platform the map. To enable the lasers to detect all the features and fluctuations of the surroundings, the KMR was operated slowly in several orientations. The speed, direction, and time spent scanning and re-passing over the surroundings all affect the quality of the map's representation of the space.

Once the mapping method has been completed, the mapping procedure can be initiated through KUKA's software package, KUKA Sunrise OS (41), which facilitates communication with the KMR. This map is then loaded onboard the KMR. To verify the robot's localization ability, the platform then navigates the surroundings using the loaded map in order for the platform to localise itself. Once content with the quality of its localization, the platform signals it has successfully found itself in the environment.

Using the features made available by the KUKA Sunrise OS, positions can be taught by letting the integrated platform's laser sensors to map the surroundings at a predetermined position. The first position to be taught is the home position. After determining the X and Y values of the home position, each additional position on the map is manually taught along with their appropriate offset. The robot then drives to the new locations and remaps at each position to validate these places ensuring it has the best probability of finding itself at each of these positions.

Both the rotational tests and the strafing tests follow the same procedure. After the positions are taught, they remain constant for all tests and speeds.

## 3    Procedure

The testing method would consist of the two previously introduced path options, strafing, and rotations. Each variation was tested ten times, with consecutive runs commencing from the previous stop, consistent with ASTM standard F3244-21 (40). This provided additional information about a final displacement that can be compared to the average displacement between runs. The positional data would be obtained from both the KMR and the Leica's tracking system. The ground truth Leica data were compared with the KMR positional data. Of the two path options, four further variations would be tested.

- A strafing test using only the encoders,

- A strafing test utilising the mapping features,

- A rotations test using only the encoders,

- and finally, a rotations test using the mapping features.

Within these four variations, 3 different speeds were tested. Three speeds were chosen to validate if there was any relation to the accuracy and repeatability of the platform with speed. The three speeds chosen for testing were, 0.08 m/s, 0.16 m/s and 0.24 m/s. These speeds were selected since they allowed for testing both slow controlled movements and faster speeds that could be conducted safely in a limited space.

Tests on encoders only considered the slowest speed since inspections are not intended to be performed solely without mapping. The speed parameters were adjusted for mapping tests, the most likely use case, while the slower speeds were compared across all four variations to achieve a baseline of comparison. The slowest speed is probably the speed at which the platform would move at during typical NDT inspection procedures. With mapping capabilities enabled, the three speeds were tested for both strafing and rotations. The tests performed are summarised in Table 1.

Table 1. Experimental Variations with speeds

| Speeds (m/s) | Tests | | | |
|---|---|---|---|---|
| | Strafing w/ Encoders | Strafing w/ Mapping | Rotations w/ Encoders | Rotations w/ Mapping |
| 0.08 | ✓ | ✓ | ✓ | ✓ |
| 0.16 | | ✓ | | ✓ |
| 0.24 | | ✓ | | ✓ |

A guideline was setup to define the results and information that have been acquired. There is currently not a standard for the definitions of capturing the accuracy and repeatability of a mobile platform so suggested definitions have been made. Firstly, Repeatability in this work has been defined as "How accurately is one able to return to ones starting position". Repeatability displacement error has been defined as the "Distance value from the initial starting position". Lastly, the repeatability displacement accuracy has been defined as "Difference between ground truth, and measured value", for this case the difference between the ground truth Leica laser tracker, and the reported KMR value.

The expectations of these experiments,

1. Strafing experiments to show the best repeatability accuracy.

2. Addition of the map should greatly reduce the error, improving repeatability.

3. Speed increases should reduce accuracy and worsen repeatability error.

## 4    Results

In this section, the findings of the study, are presented. The results are organized into subsections, starting with the positional graphs, highlighting the starting positions of the platform for each run. This is followed by the accuracy and error analysis for the different test scenarios. Key findings related to the research question are highlighted throughout the section.

### 4.1    Positional Data

The starting positions of the KMR platform for every run was recorded for each test. This data was recorded with both the KMR and the ground truth Leica. A visual representation of any trends or consensus among results can be obtained by plotting the precise positions. The trends are presented below, for both the encoder tests and first mapped tests. The scales for the Leica and KMR are in their respective formats from the measured positions and are presented to show the overall trends.

### 4.1.1  Test Path: Strafing

Figure 5 illustrates the return to start variations observed in the strafing tests. Figure 5 a) and b) show the use of the encoders only, strafing at a speed of 0.08 m/s.

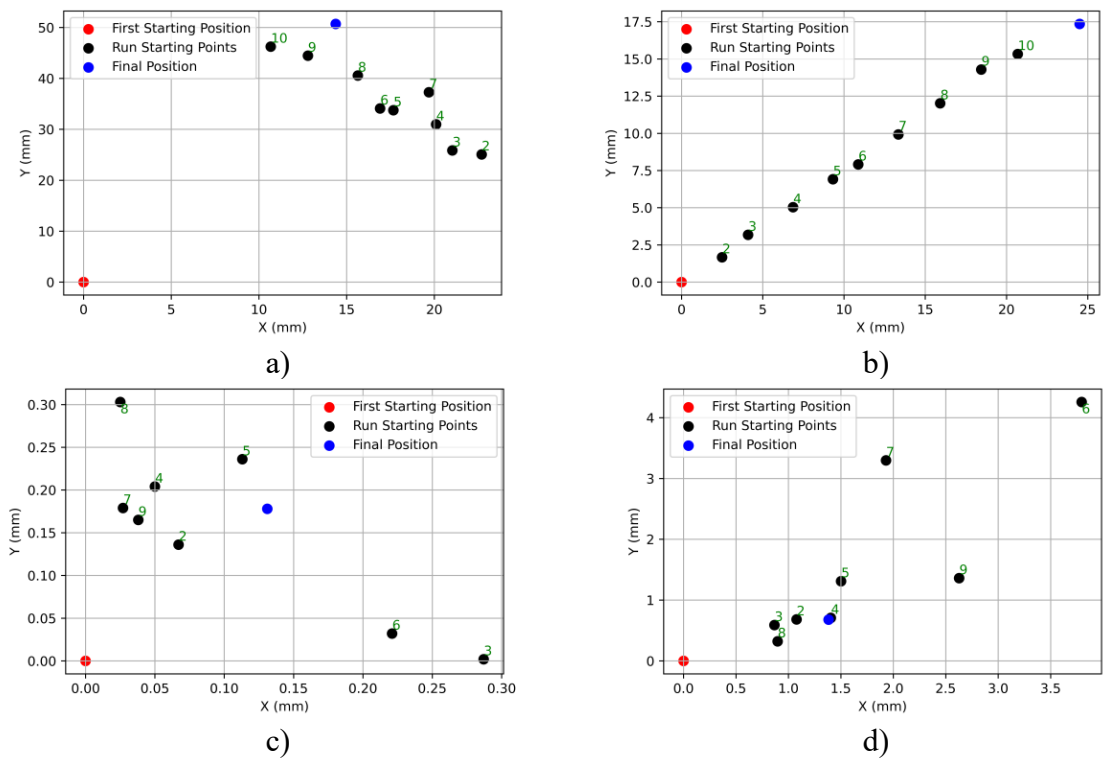

Figure 5. Return to start positions for Strafing tests at 0.08 m/s. Encoder: a) KMR, b) Leica, Map: c) KMR, d) Leica.

What can be observed from Figure 5 a) and b) is that the KMR reports spread out positions without any structure. According to the Leica, there is a continuous offset in each cycle, with the starting point seemingly following an overall trend. Figure 5 c) and d) shows the return to home positions with the mapping features enabled.

The KMR readings show a scale in which the observed variations with spreads of 0.3 mm in x and y. On the contrary, the Leica measures a greater variation that was over 3.5 mm each. Based on the observed data points, it appears that the KMR tends to overestimate its accuracy capabilities when the mapping features are activated. Unlike what was observed in the strafing with encoders in Figure 5 a) and b), there is no clear pattern in the reported positions, and the scale of variation is also greatly reduced.

The same trend is observed for the two other test speeds as was seen for the 0.08 m/s mapped test. The KMR once again reported smaller variations with spreads of positions of < 0.5 mm, which is refuted by the measurements made by the Leica, suggesting larger spreads and variations between the points, measured at > 5 mm. As was with the 0.08 m/s mapped test, the two larger speeds, 0.16 m/s and 0.24 m/s, also did not display any notable pattern. All three mapped tests had spreads of data smaller than the encoder test.

### 4.1.2 Test Path: Rotational

The trend is this time presented by both the KMR and the Leica system for the rotational tests, displayed in Figure 6.

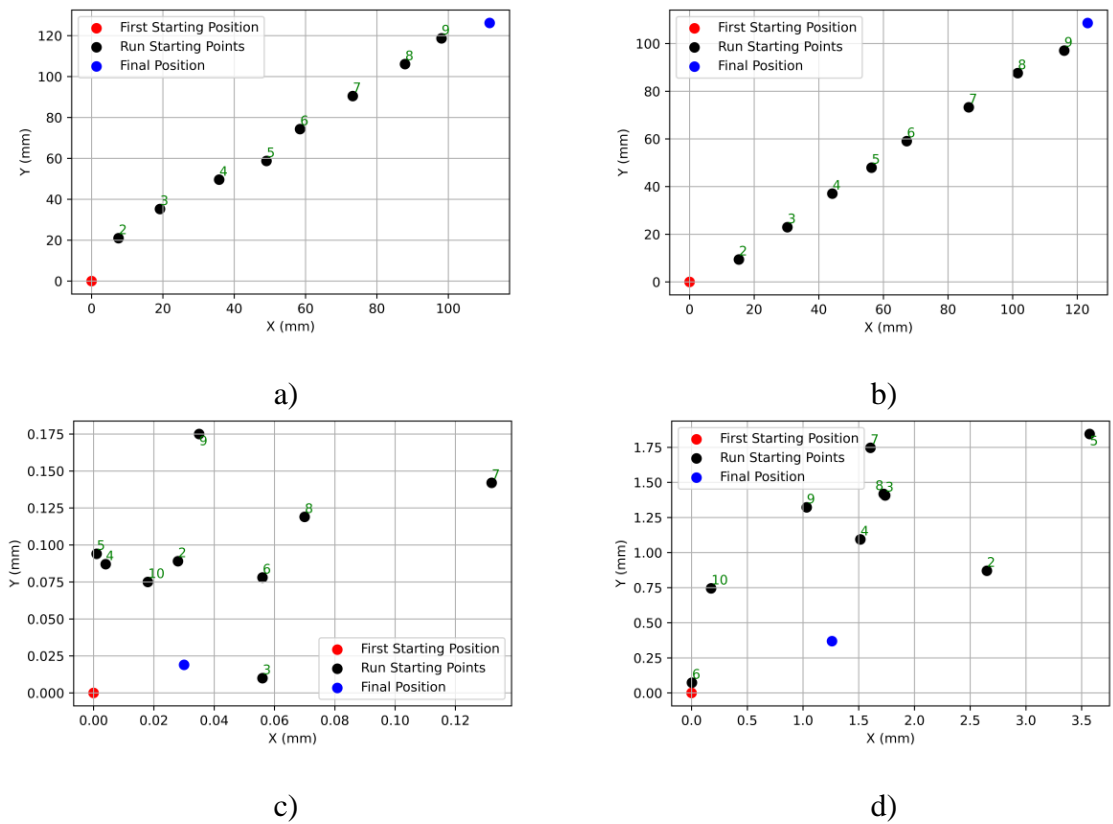

Figure 6. Return to start positions for Rotational tests at 0.08 m/s. Encoder: a) KMR, b) Leica, Map: c) KMR, d) Leica.

For the encoder test, Figure 6 a) and b), the overall displacement from the introduction of rotations far surpasses the values seen with the strafing, with a spread in data over 75% larger. The differences in the x and y components being over 100 mm after all the runs for the rotational test. The KMR platform seems to have a consistent displacement error when using only the encoders.

When introducing mapping features, the displacement error of the mobile manipulator improves by over 90%. This is shown in the positions measured in Figure 6 c) and d). As previously reported in the strafing tests with mapping enabled, the KMR inaccurately estimates the true x and y positions during the rotational mapping. The KMR once again suggests sub millimetre final displacements with the Leica reporting values closer to 3.5 mm. This trend is once again observed for both the higher speeds.

## 4.2 Repeatability Error and Accuracy

This section explores the final and average repeatability error and accuracy as measured by the Leica and the KMR. As previously defined in Section: 0, the error measured is as found with the Leica system. The accuracy is the difference between the values reported by the KMR and the Leica.

### 4.2.1 Strafing

The measurements of the errors and accuracies have been analyzed with the respective X and Y components as defined in Figure 2. The average accuracy comes from taking the average displacement reported from the ten runs, for both components. These values are reported by the KMR and Leica, with the difference between both being how accurate the KMR reports its displacements. The error is simply the average displacement measured by the Leica across the ten runs. The components of the average displacements and displacement errors are shown in Figure 7 a) and b).

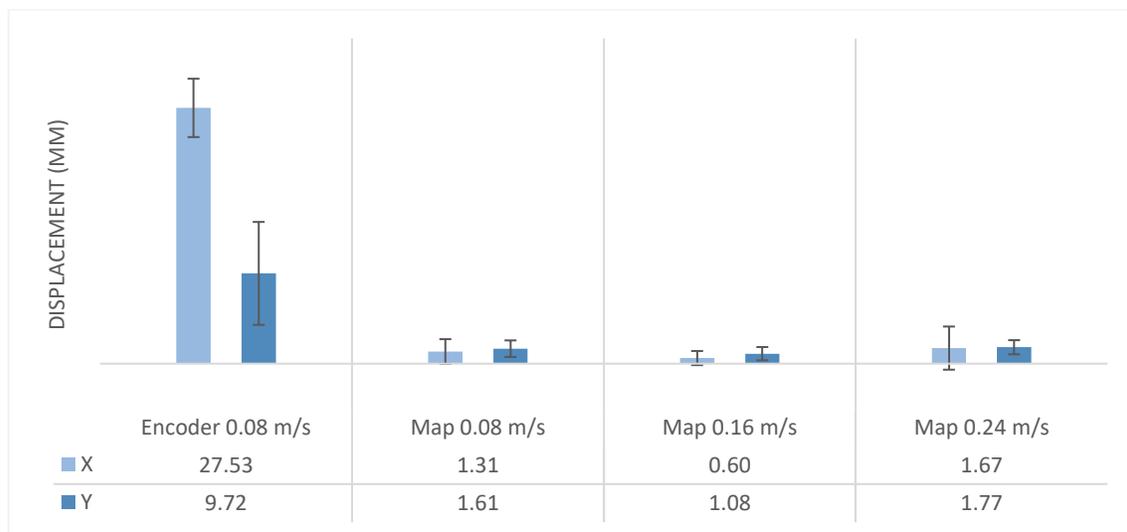

| | Encoder 0.08 m/s | Map 0.08 m/s | Map 0.16 m/s | Map 0.24 m/s |
|---|---|---|---|---|
| X | 27.53 | 1.31 | 0.60 | 1.67 |
| Y | 9.72 | 1.61 | 1.08 | 1.77 |

a)

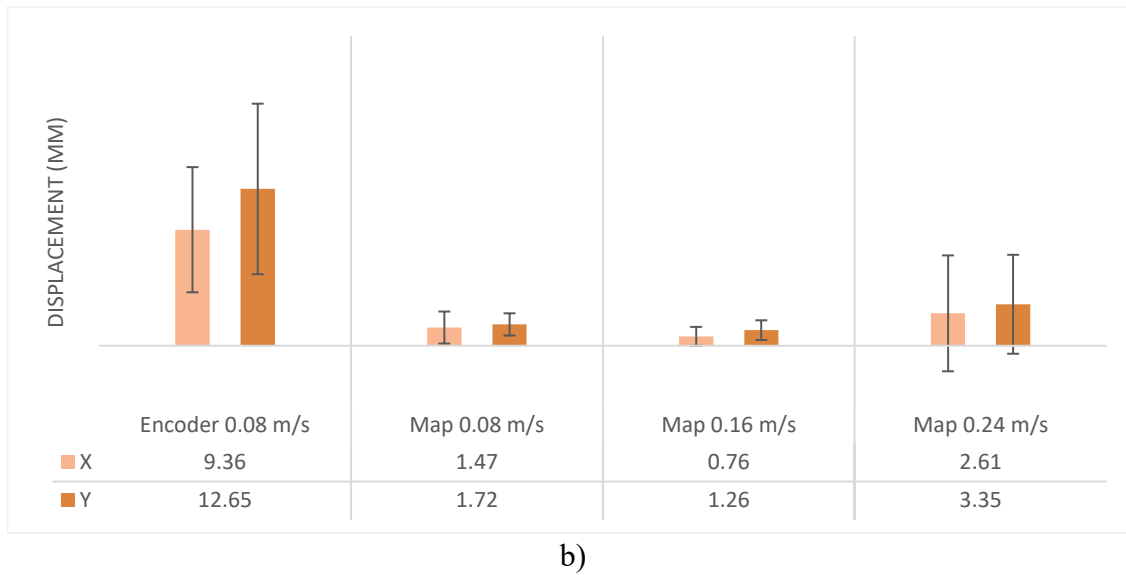

| | Encoder 0.08 m/s | Map 0.08 m/s | Map 0.16 m/s | Map 0.24 m/s |
|---|---|---|---|---|
| X | 9.36 | 1.47 | 0.76 | 2.61 |
| Y | 12.65 | 1.72 | 1.26 | 3.35 |

b)

Figure 7. Average strafing displacement results a) Average strafing accuracy components. b) Average strafing error components.

By enabling of the mapping features, between the two 0.08 m/s tests, an improvement was observed in both the X and Y components across both the accuracy and error. A notable improvement of 95.2% was found in the X component accuracy.

The accuracy and error observed are much improved across the three mapping speeds in comparison to the use of purely the encoders. Consistent results of sub 2 mm in the accuracy across both the X and Y components are observed, with the 0.16 m/s test displaying the best accuracy components. The errors, in Figure 7 b), perform similarly to the accuracies for the 0.08 m/s and 0.16 m/s mapping tests, with a marginal increase observed for the greatest speed of 0.24 m/s. The errors are slightly larger in the 0.24 m/s, suggesting the KMR platform not able to return to the same position as well as at the lower speeds.

Y components tend to perform slightly worse than X in mapping tests, which is consistent with the Omnidirectional wheel operation. When the movement is in the Y direction, the Mecanum wheels on each side move in opposite directions, which can lead to wheel slipping or friction adversely impacting the Y movements. The X component's accuracy is significantly worse when the encoder is turned on, suggesting that the platform perceives itself to be performing better than it does. The encoder's error shows a consistent pattern across all mapping tests, with the Y component performing worse than the X component.

### 4.2.2 Rotational

The results of the rotational tests are in the same format as those in the strafing, Section 4.2.1. The resultant averages of the rotational tests are presented in Figure 8. Broken down into the components of the averages and errors in Figure 8 a) and b).

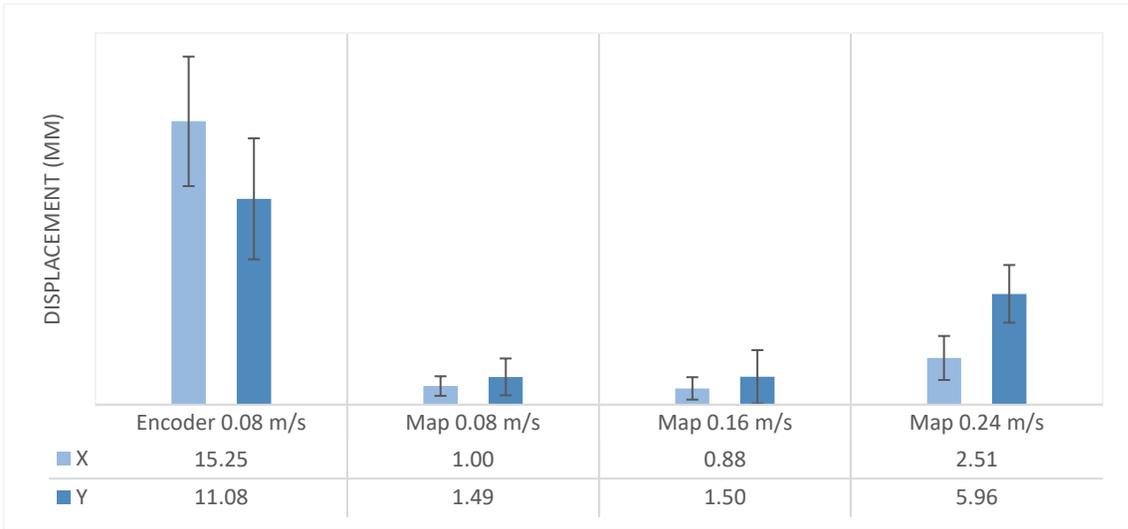

a)

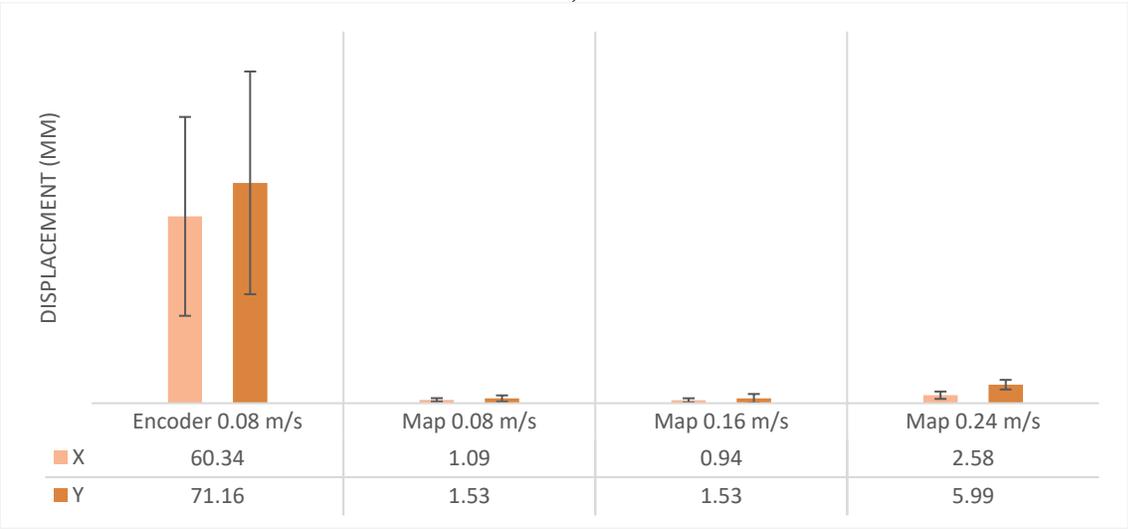

b)

Figure 8. Average rotational test displacement results a) Average rotational test accuracy components. b) Average rotational test error components.

The first observation that can be made with Figure 8 b) is the larger average error seen with the use of the encoder. A resultant error of 93.29 mm, from the x and y components, which is undesirable for any application scenario. However, by utilizing the mapping features the resultant rotational test error is improved by 98% while the accuracy also sees an improvement of over 90%. The mapping features make a remarkable difference especially with the introduction of the more complicated movements.

Once again, across the mapping speeds of 0.08 and 0.16 m/s, the average accuracy and errors observed are similar across both X and Y components. But the largest speed of 0.24 m/s sees drop offs in the accuracy and an increase in error, especially in the Y component.

### 4.2.3 Standard Deviation and Variation

The standard deviations of the data across all the tests are shown in Table 2, accompanied by the maximum difference in both X and Y components in each of the respective tests. Sigma (σ) represents the standard deviation, with Delta (Δ) representing variation.

Table 2. Standard deviation and max variation, a) Strafing tests, b) Rotational tests

| | σ X (mm) | σ Y (mm) | Max Δ X (mm) | Max Δ Y (mm) |
|---|---|---|---|---|
| **Encoder 0.08 m/s** | 5.79 | 7.87 | 17.35 | 24.50 |
| **Map 0.08 m/s** | 1.56 | 1.87 | 4.97 | 5.73 |
| **Map 0.16 m/s** | 1.07 | 1.30 | 3.72 | 4.26 |
| **Map 0.24 m/s** | 4.75 | 4.58 | 16.62 | 17.39 |

a)

| | σ X (mm) | σ Y (mm) | Max Δ X (mm) | Max Δ Y (mm) |
|---|---|---|---|---|
| **Encoder 0.08 m/s** | 37.35 | 42.41 | 108.64 | 123.21 |
| **Map 0.08 m/s** | 1.22 | 1.46 | 3.59 | 5.30 |
| **Map 0.16 m/s** | 0.82 | 2.00 | 2.88 | 7.14 |
| **Map 0.24 m/s** | 1.42 | 2.38 | 4.75 | 8.46 |

b)

In the strafing tests, Table 2 a), the deviation in the encoder data shows a larger spread amongst the Encoder test in comparison to the mapping 0.08 m/s and 0.16 m/s tests. The max difference in X and Y for the 2 slower mapping speeds is relatively small compared to the speed of 0.24 m/s where a greater spread is observed. Once again across all the data in Table 2 a), the Y tends to preform worse than the X component.

Like the strafing test results, the deviation in the encoder data of the rotational tests, Table 2 b), show a larger spread amongst the Encoder test in comparison to the mapping 0.08 m/s and 0.16 m/s tests. Across all the data in Table 2 b), the Y component tends to preform worse than the X. The addition of the rotation has added a greater spread in the data with the max difference in the X and Y components being significantly larger than the previous mapped tests.

## 4.3   Average Path Distances

The previous sections described the positional tests, investigating the spread of the data regarding returning to the starting positions, dealing purely with the repeatability. With the tests comprising of four 1-meter paths, it is also possible to analyse the performance of the KMR platform at travelling the correct specified distance. The Leica is used as the ground truth to report the true distance travelled. The error is the difference between the reported KMR value and the Leica.

### 4.3.1 Strafing Path Distances

The path distances from the strafing tests are broken into the X and Y components, due to the different nature of movement required for each. These are average path lengths calculated from each stage across all 10 runs. The average path distances are presented in Figure 9.

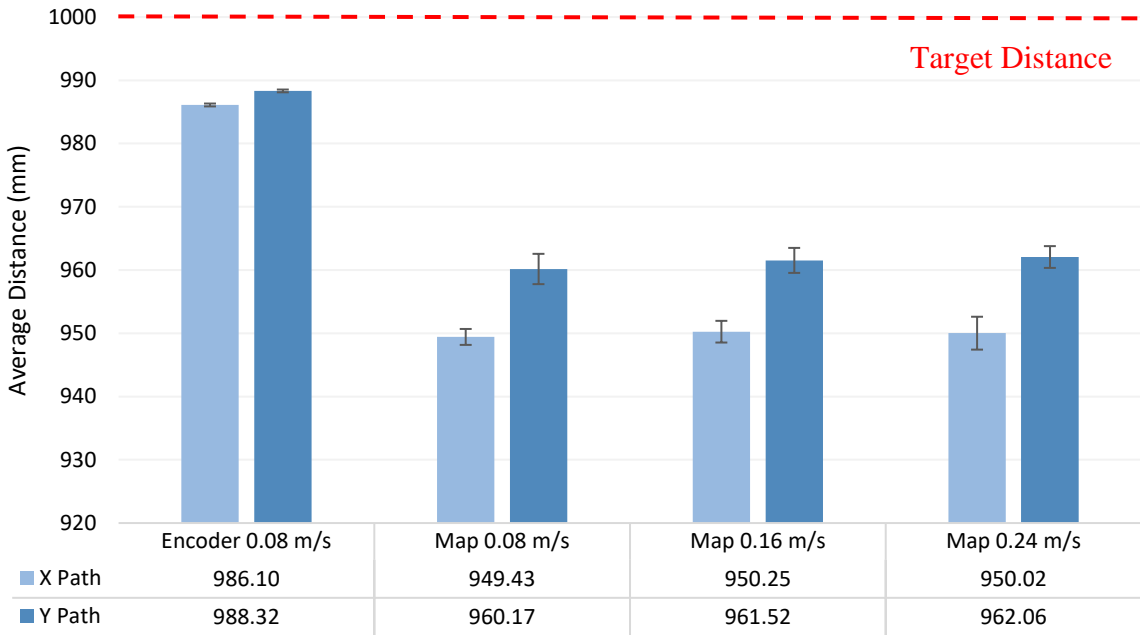

| | Encoder 0.08 m/s | Map 0.08 m/s | Map 0.16 m/s | Map 0.24 m/s |
|---|---|---|---|---|
| X Path | 986.10 | 949.43 | 950.25 | 950.02 |
| Y Path | 988.32 | 960.17 | 961.52 | 962.06 |

Figure 9. Strafing average path distances (error bars represent standard deviation)

From the values displayed in Figure 9 it can be seen that the mobile manipulator on average does not reach the complete target distance of 1000 mm. For optimal results, tests should have all paths measuring 1000 mm. When operating with the encoders. the KMR does travel close to the target, but undershoots by 10 – 15 mm. The expectation with the mapping results was to observe a more constant distance travelled across all runs with values closer to the target 1000 mm.

The first observation is that the map enabled data has smaller path distances travelled in comparison to that observed in the encoder paths. These paths are also seen to be consistent across the three speeds. The Y paths are longer than the X paths, displayed across all tests, and more noticeably with the mapped tests. The standard deviation is small across all paths, with a maximum value of ±2.6 mm, indicating the KMR is consistent in the path distance travelled, but less so than the encoders.

Comparing the values of the KMR and ground truth gives the measured error, which is displayed in Figure 10.

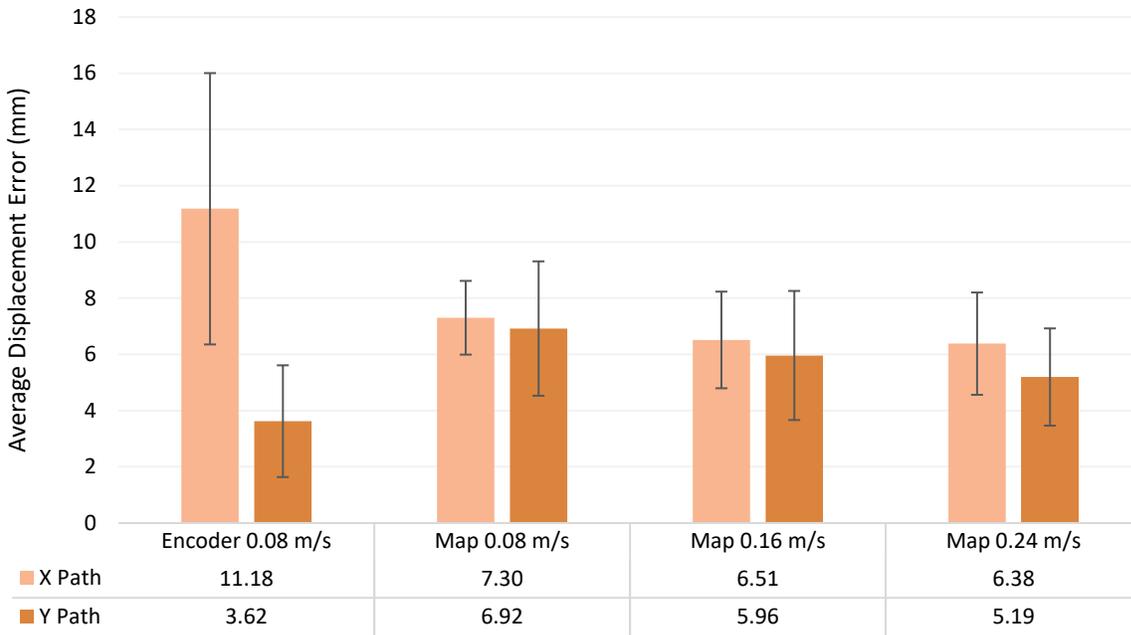

| | Encoder 0.08 m/s | Map 0.08 m/s | Map 0.16 m/s | Map 0.24 m/s |
|---|---|---|---|---|
| X Path | 11.18 | 7.30 | 6.51 | 6.38 |
| Y Path | 3.62 | 6.92 | 5.96 | 5.19 |

Figure 10. Measured error of strafing path distances

The mapped tests display consistent errors, even showing a slight trend of improved error with the increase in speed. Although the distance travelled was smaller with the X components, the error reported is slightly larger than its Y counterpart.

### 4.3.2 Rotational Path Distances

Since the rotational test always faced in the X direction, and hence, it only had one component. The average path distance travelled with the mapping enabled at 0.08m/s is shown in Figure 11.

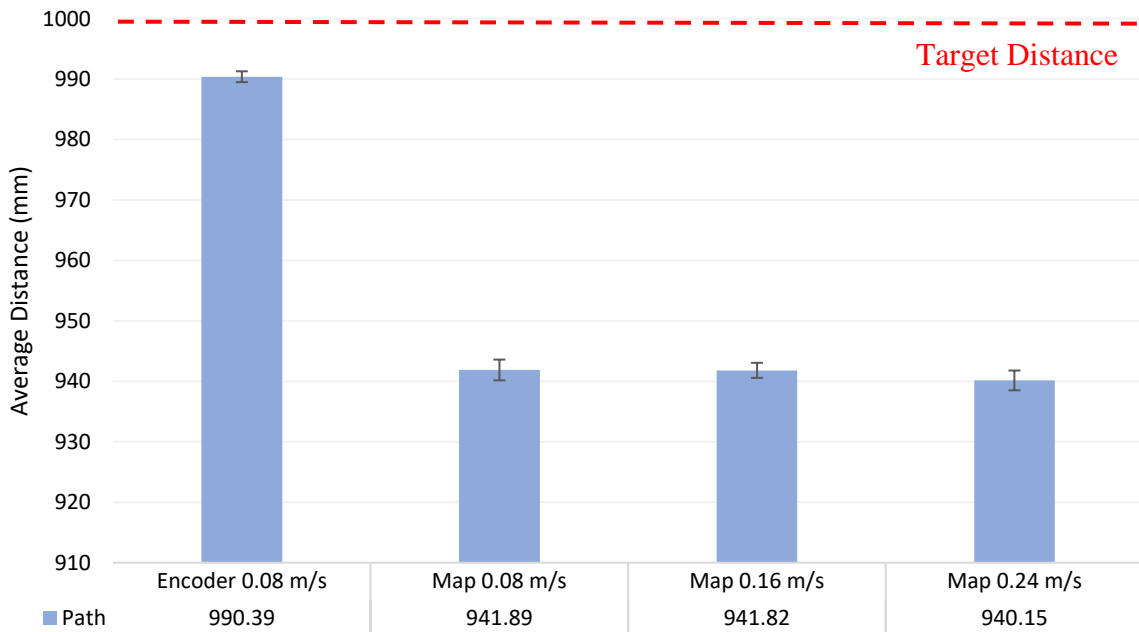

| | Encoder 0.08 m/s | Map 0.08 m/s | Map 0.16 m/s | Map 0.24 m/s |
|---|---|---|---|---|
| Path | 990.39 | 941.89 | 941.82 | 940.15 |

Figure 11. Rotational average path distances (error bars represent standard deviation)

The first thing to note is that the data presented in Figure 11 have smaller path distances travelled in comparison to that observed in the strafing tests in Figure 9. These mapped paths are once again similar across all three speeds. The encoder test shows an averaged travelled distance of 990 mm, the closest to the target distance across all tests. The standard deviation is small across all paths, so the KMR seems to consistently undershoot its target distance.

The error observed in the rotational tests however is larger than the strafing tests, displayed in Figure 12.

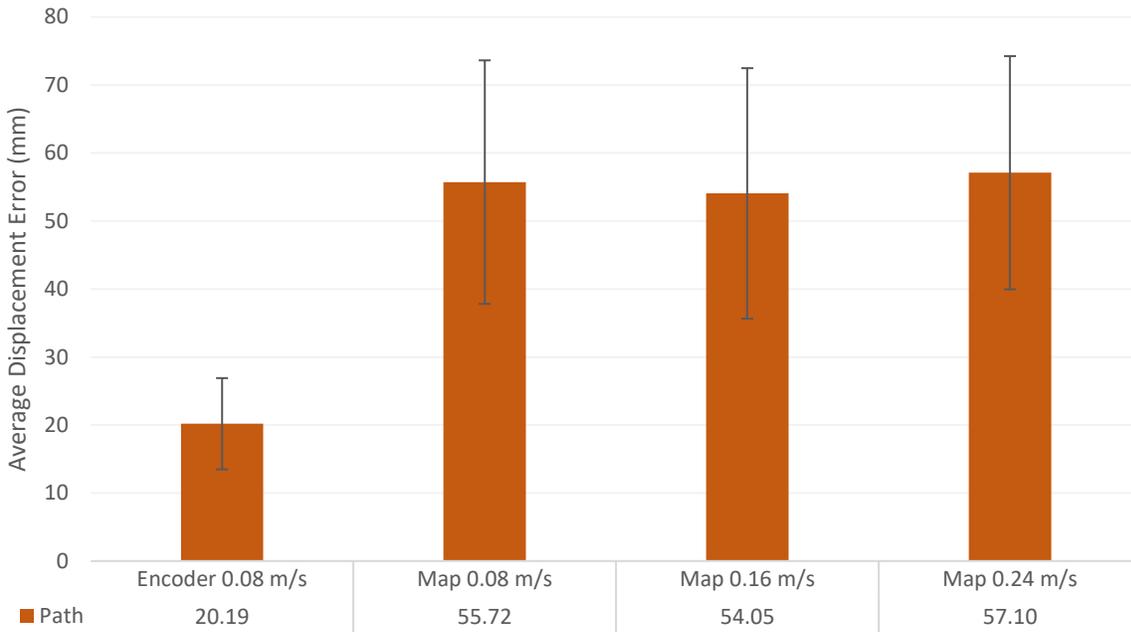

Figure 12. Measured error of rotational path distances

The errors observed in Figure 12 are significantly larger than the strafing tests, with errors of over 50 mm for each of the mapped speeds. The encoder test shows the smallest error but is still a greater value compared to the strafing result. The increased error means the KMR is not as capable of reporting its true travelled distance with the introduction of the complicated motions.

## 4.4    Path Angle Deviations

The path deviations were extracted from the 1-meter sections that made up the full test paths. The deviations are presented in their individual sections for each of the tests. The graphs are presented in the supplementary material, Supplementary Figure 1, and Supplementary Figure 2.

The strafing test path deviations are recorded in Supplementary Figure 1. The mean angles, in Supplementary Figure 1 a), a trend can be observed of an increase in speed correlating with an increase in angle deviation. The mean path deviations are also seen to increase for each of the paths, with an increase of path deviations of 1 degree between paths 2 to 4. The encoder test can be seen to perform the best with the least amount of deviation, with a max deviation of 0.35 mm observed on the $3^{rd}$ path.

The trend observed in the mean angle deviation is not reflected in the error, however. The error remains consistent across the mapped tests for the strafing tests, with all tests under 0.4 degree reported error. The encoder tests show a larger error in paths 2 and 4, which correspond to the motion in which the platform is moving horizontally.

The rotating path mean angle deviation, in Supplementary Figure 2 a), once again displays the encoder test having the smallest degree of deviation, with a max deviation of 1.15 degrees measured. The trend of increased speed correlating to increased deviation is not observed in these results, and the deviation is seen to be larger than the strafing tests with max deviations reaching 3.5 degrees.

The error for the rotating path tests, Supplementary Figure 2 b), does not present a consistent trend, but the encoder error is observed to be smaller than the mapped tests across all paths.

## 4.5   Mean Test Path Error

To further quantify and understand the performance of the KMR platform, the mean test path error has been extracted. Due to slight differences in packets received between the KMR and the Leica, the KMR data has been interpolated, to allow direct comparison to the ground truth Leica at every data point captured across all tests. The nature of the rotation stage meant that a direct interpolated comparison was not possible, but the strafing data has been extracted and is presented in Supplementary Figures 3-6.

The encoder test, Supplementary Figure 3, presented larger errors in the first and last paths, the largest error being recorded in the last path. This error reached a max recorded error of 22 mm. The mapped 0.08 m/s test, Supplementary Figure 4, showed error increasing over the whole test path rather than simply the first and last path. The last path once again also presented the largest spike in error, with a max error of 15.6 mm.

For speeds 0.16 and 0.24 m/s the errors appear similar, with distinct spikes across the last three paths. Once again, the final paths display the largest errors, 19.2 mm and 24.5 mm respectively.

# 5    Discussion

Understanding the repeatability, error and accuracies associated with the KMR mobile robotic platform are critical to planning and operation for NDT purposes. Therefore, characterising the displacement accuracy and displacement error of each variation was vital. The accuracy was quantified by finding the difference between the reported KMR data, and the experimental ground truth, from the Leica Laser Tracker. The displacement error was the reported Leica value, due to its high degree of precision. The measurements of the errors and accuracies were analyzed with the respective X and Y components, as defined in Figure 2.

The KMR platform seems to have a consistent displacement error when using only the encoders observed across both encoder tests, Figure 5 (a, b) and Figure 6 (a, b). The final displacement with the rotational tests is much larger than the strafing when using only the encoders. A value of over 100 mm is seen with the rotations compared to the under 40 mm with the strafing. Suggesting that the more complicated paths have a direct correlation with a greater error. The KMR seems to under-report the true X and Y positions in the mapping tests, by reporting sub-millimetre spreads for the positions, compared to ground truth reporting values of around 5 mm across all mapping tests and speeds.

The final positional data that was recorded and reported in Section 4.1 was summarised graphically in the repeatability error and accuracy results. The repeatability error measured across all tests indicated a consistent pattern in which the error observed in the Y is worse than the X component, regardless of positional accuracy. This is something which is consistent with the omnidirectional wheel operation. During strafing operations, the X component of the path involved moving all the wheels forward in unison. In contrast, the Y component required moving a front and back wheel forward while the other two wheels moved backward in order to slide across effectively. This driving technique is more prone to errors due to factors like friction, wheel spin, and the complexity of the driving action, and was reflected in the experimental results.

The accuracy observed across all the tests highlighted the KMR's tendency to perform worse in the X component for the encoder runs. This may be due to suboptimal calibration of the encoders. Excessive vibrations can also lead to errors, which may be affected by driving speeds and uneven flooring. Mapping features improved both the accuracies and errors observed. The resultant strafing error improved by over 85%, with an improvement in accuracy of over 92%. Rotational tests resultant errors were improved by 98%, alongside resultant accuracy improvements of over 90%.

The strafing accuracies performed well across the mapped tests, Figure 7 b), notably all smaller than 2 mm. The rotational test saw sub 2 mm for the two lower mapped speeds, but the highest speed saw the accuracy components rise to 2.51 mm and 5.96 mm for X and Y, respectively.

The spread of data observed across all tests was documented in Table 2. The spread in Y components was larger by an average of 20% than in X components. For the mapped tests, the data spread was much smaller than the encoder at all stages. The standard deviation observed was comparable between strafing and rotational mapped tests at the two lower speeds. The 0.24 m/s saw a greater spread in data for the strafing test, but this was due to an outlier during the 7th run. The rotational encoder test standard deviation was elevated compared to the strafing test, with the max variance in X and Y components measured at over 100 mm. These highlight that the introduction of the complex rotation affects the repeatability of the KMR when using the encoders. Despite the decreased performance of the encoders with the introduction of a more complex path, the rotational tests showed very similar results to the strafing tests when mapping features were enabled for both displacement error and accuracy. This emphasizes the significant and immediate improvement in localization performance that came from using the laser features.

Another factor examined in this paper was evaluating the mobile manipulator's ability to traverse the intended distance. The target distance was 1000 mm for each of the four paths that formed a run. For both the rotational and strafing tests, the average path length fell short of the target of 1000 mm. The encoder distance travelled was closer to the target distance than the mapped tests. The KMR falling short by 12 -15 mm for the strafing and 10 mm with the rotational tests when using the encoders. Compared to the difference of over 40 mm of each path with the mapping enabled. The Y paths were longer than the X paths, displayed across all tests, and more noticeably with the mapped tests. The path lengths were consistent across the three mapped speeds, a maximum standard deviation of $\pm 2.6$ mm measured, indicating the KMR was consistent in the path distance travelled.

The errors, however, displayed that the X paths were shorter, and displayed the larger error values, shown in Figure 10. The mapped tests displayed consistent errors, with a slight trend of reduction in error observed as the speed increased. The mapped rotational tests showed that the three speeds were once again consistent in their reported path distance travelled, once again with a standard deviation of less than $\pm 1.8$ mm. The mapped error in this case performed worse than the encoder, with a much larger value of over 50 mm at each speed.

Breaking down the X and Y components into their individual paths, the errors observed during the strafing tests remained consistent across all paths and speeds in the mapping trials. Conversely, in rotational tests, the error notably increased by over 35 mm per path, from the original average of 7 mm. Interestingly, a shift in trends occurred, with the first path exhibiting a greater error compared to paths 2 and 3. Notably, across all three mapping trials, the fourth path consistently demonstrated the highest error, averaging 86 mm. Furthermore, examination of the distance covered for the fourth path revealed it consistently as the shortest travelled path during the mapped rotational tests.

The discrepancy between the shorter path distance and the observed large error indicates a potential overestimation by the platform regarding the actual distance travelled along the fourth path. The underlying cause for consistently failing to reach the target distance is not immediately apparent and requires further investigation.

The path angle deviations during the strafing tests showed sub-2-millimetre values, with higher speeds correlating to larger deviations. Notably, the encoder outperformed other mapped tests, recording a maximum deviation of 0.35 degrees, while the three other paths showed deviations better than 0.1 degrees. Similar results were observed in the rotational tests, with a maximum deviation of 1.15 degrees and other paths performing better than 1 degree deviation. Interestingly, after the first path, there was a significant decrease in performance deviation in the mapped tests, with a measured change of over 1 degree between the first and second paths. The reason for the higher deviation in the mapped tests compared to the encoder in both strafing and rotational tests is unclear, and there seems to be no correlation between final displacement accuracy and deviation angles.

In Supplementary Figures 3-6, the mean error of the complete test paths was measured. It was observed that errors tended to increase as the speed of the platform increased, with all tests showing a larger error in the last path. The spikes in error were found to correspond with the speed of the platform, with reductions occurring as the platform slowed to transition to the next stage. Interestingly, despite initially high error measurements, the error decreased significantly as the final paths concluded. The final error measured was below 3 mm for all the mapped tests and 5 mm for the encoder test.

# 6    Conclusions

The desire to explore greater flexibility in the high-value manufacturing industry by introducing mobile manipulators is a promising opportunity. For a successful introduction of such technologies, procedures and knowledge must be in place to understand the capabilities and limitations of these mobile manipulator platforms. The KUKA LBR iiwa manipulator has been previously investigated and proven to have a repeatability of <0.1 mm (28). However, there is a lack of quantification for mobile manipulator platforms like the KUKA KMR iiwa in high-value manufacturing. Utilizing mobile manipulator platforms in this sector offers significant advantages and potential. Motivated by the strict accuracy requirements for NDT applications, it became crucial to investigate the capabilities of mobile manipulator platforms.

To address this gap, a novel experimental procedure was presented in this paper to quantify and understand the limitations and capabilities of the KUKA KMR iiwa. Addressing the preliminary experimental predictions outlined in Section 3:

1. The accuracy of strafing was very similar to the rotational tests when using the mapping features. When using the encoders only, the strafing error performed better than the rotational by 80%.

2. The experimental results showed that enabling mapping significantly improved the performance of the KMR, with accuracy improvements of over 90% in both strafing and rotational tests.

3. The general trend of the error increasing with speed was observed when comparing the 0.08 m/s tests with the 0.24 m/s tests.

The positional data indicated that the KMR exhibited an underestimation of its performance when relying solely on encoders, while occasionally demonstrating a significant overestimation of its performance when mapping was enabled. The use of the encoders for complicated motions would not be recommended due to the very large errors, measured at 93.3 mm. In terms of path distance measurements, the encoders were much closer to the target distance compared to the mapped tests. The difference between intended and actual distance travelled was observed to be almost 50 mm, a finding that was unexpected.

The maximum repeatability accuracy for rotations with mapping enabled was under 7 mm. Therefore, for any inspection use case, the worst-case repeatability would be smaller than 7 mm, regardless of complex driving modes. The more realistic use case would involve driving at lower speeds, which would result in an accuracy of better than 2.5 mm and an error of less than 2.5 mm. Although the required accuracy for NDT inspections is 0.5 mm, the combination of the base and manipulator allows for the base to reach a suitable position, with the manipulator making the final fine adjustments with its more precise movements.

In future studies, we aim to expand upon these findings by evaluating how the integration of additional odometric sensing features improves the real-time precision necessary for the KMR platform. Subsequent research might entail testing in a larger environment, which could yield slightly varied results and potentially offer findings that are more representative. Additionally, upcoming projects can focus on analyzing and understanding the errors associated with moving to a designated position and then operating the arm, allowing for a more in-depth comprehension of the combined errors in mobile manipulation.

**Conflict of Interest**

*The authors declare that the research was conducted in the absence of any commercial or financial relationships that could be construed as a potential conflict of interest.*

**Funding**

This work was supported through Spirit AeroSystems/Royal Academy of Engineering Research Chair for In-Process Non-Destructive Testing of Composites, RCSRF 1920/10/32.

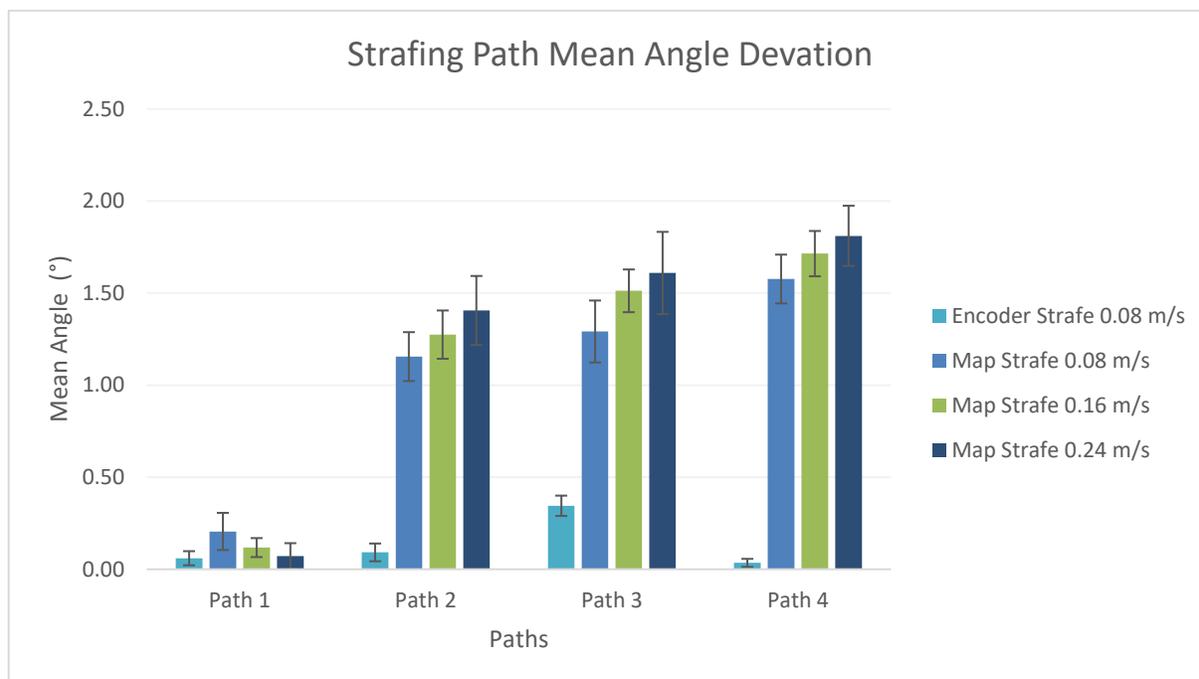

a)

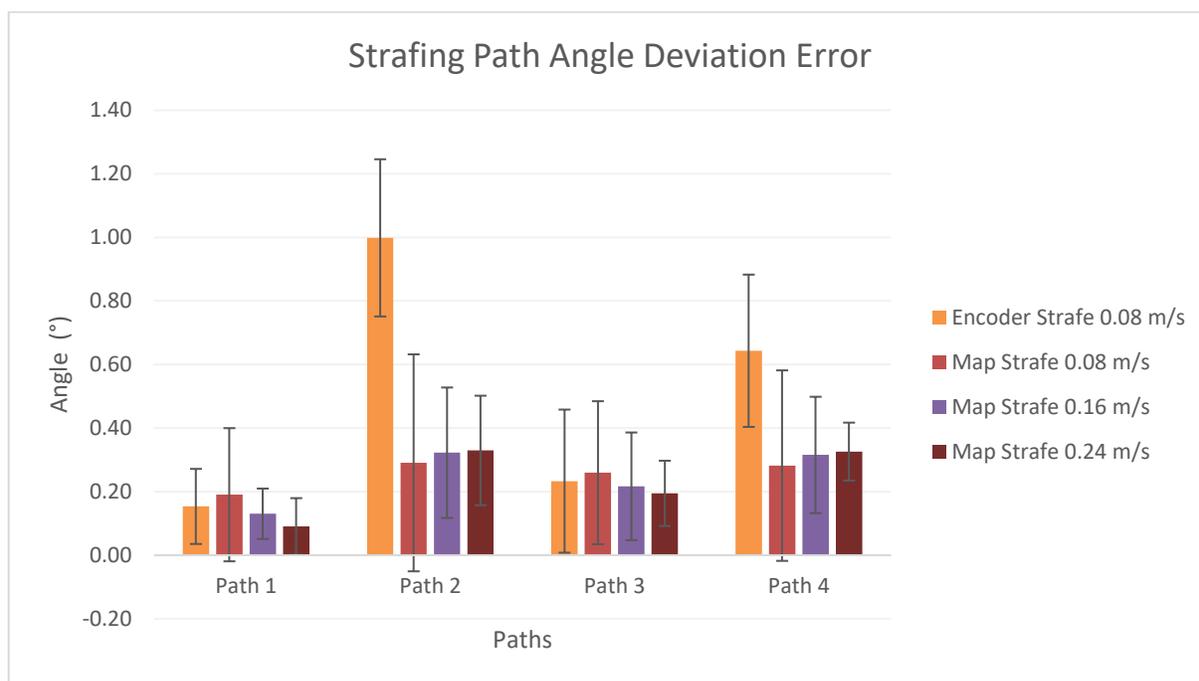

b)

Supplementary Figure 1. Strafing tests path deviations. a) Mean deviations, b) Deviation errors.

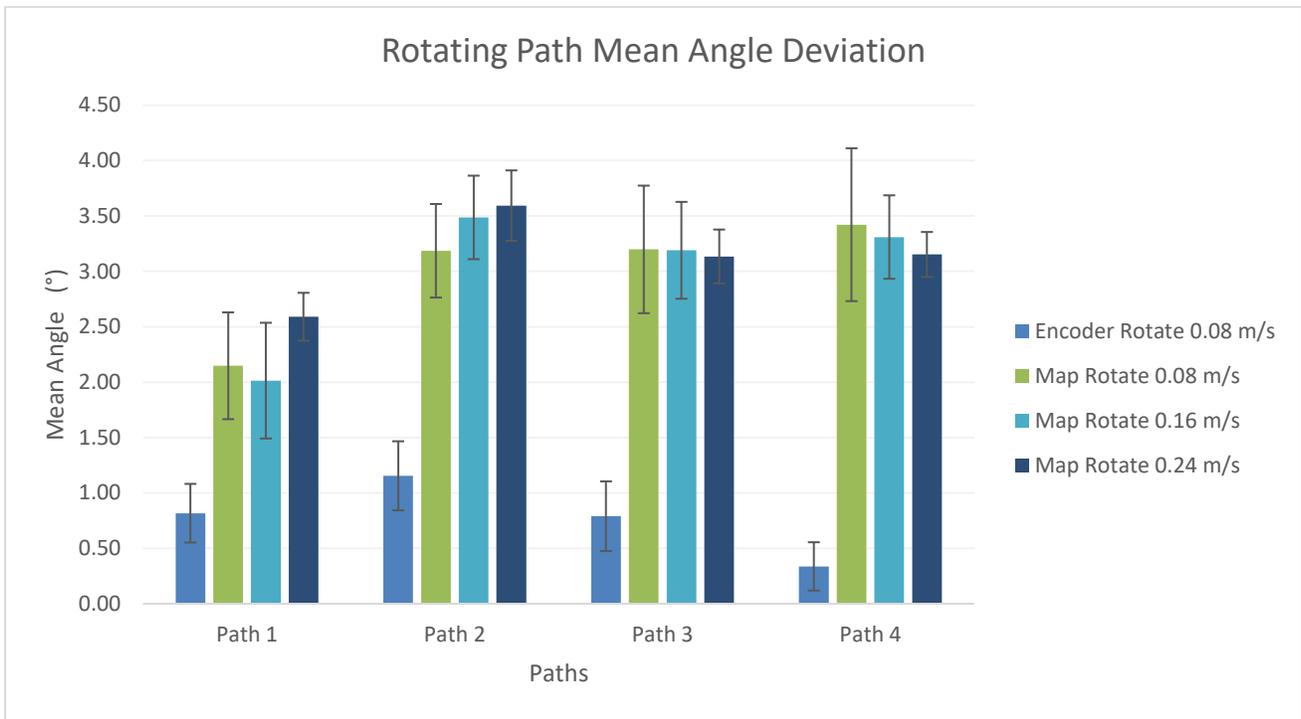

a)

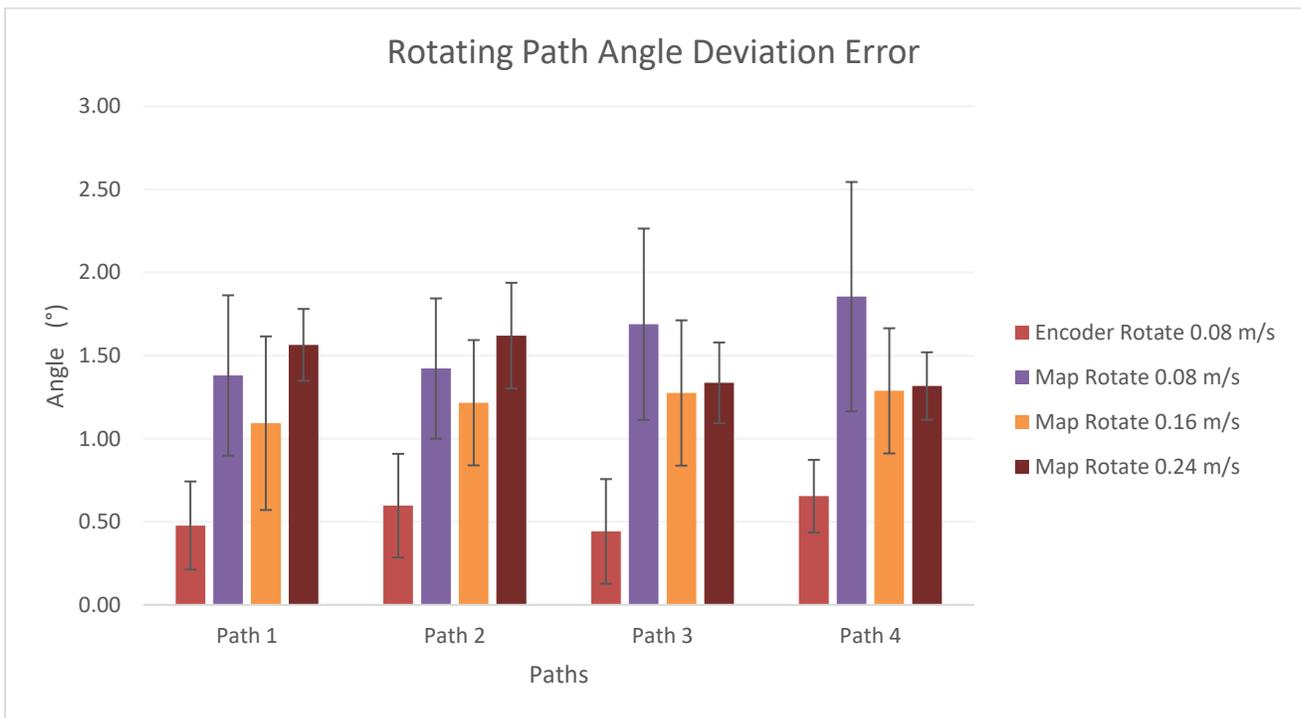

b)

Supplementary Figure 2. Rotational tests path deviations. a) Mean deviations, b) Deviation errors.

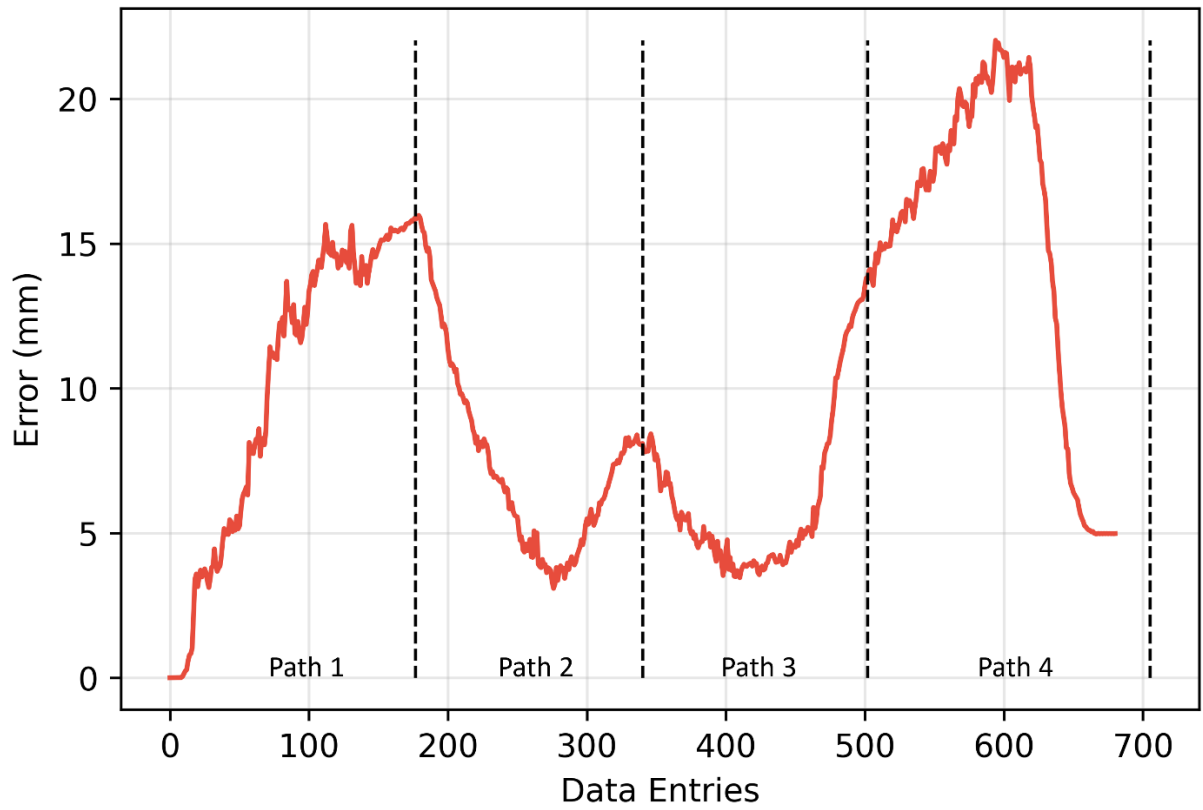

Supplementary Figure 3. Mean test error for encoder 0.08m/s.

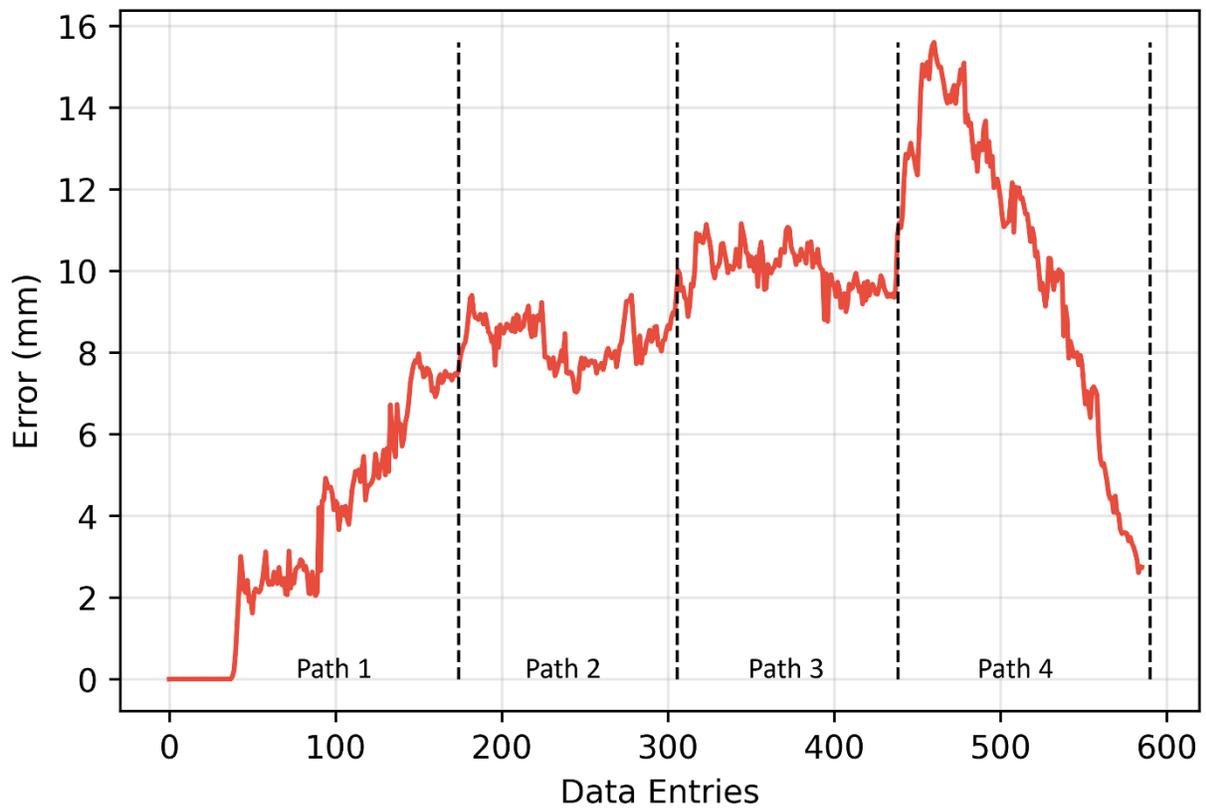

Supplementary Figure 4. Mean test error for mapped 0.08m/s.

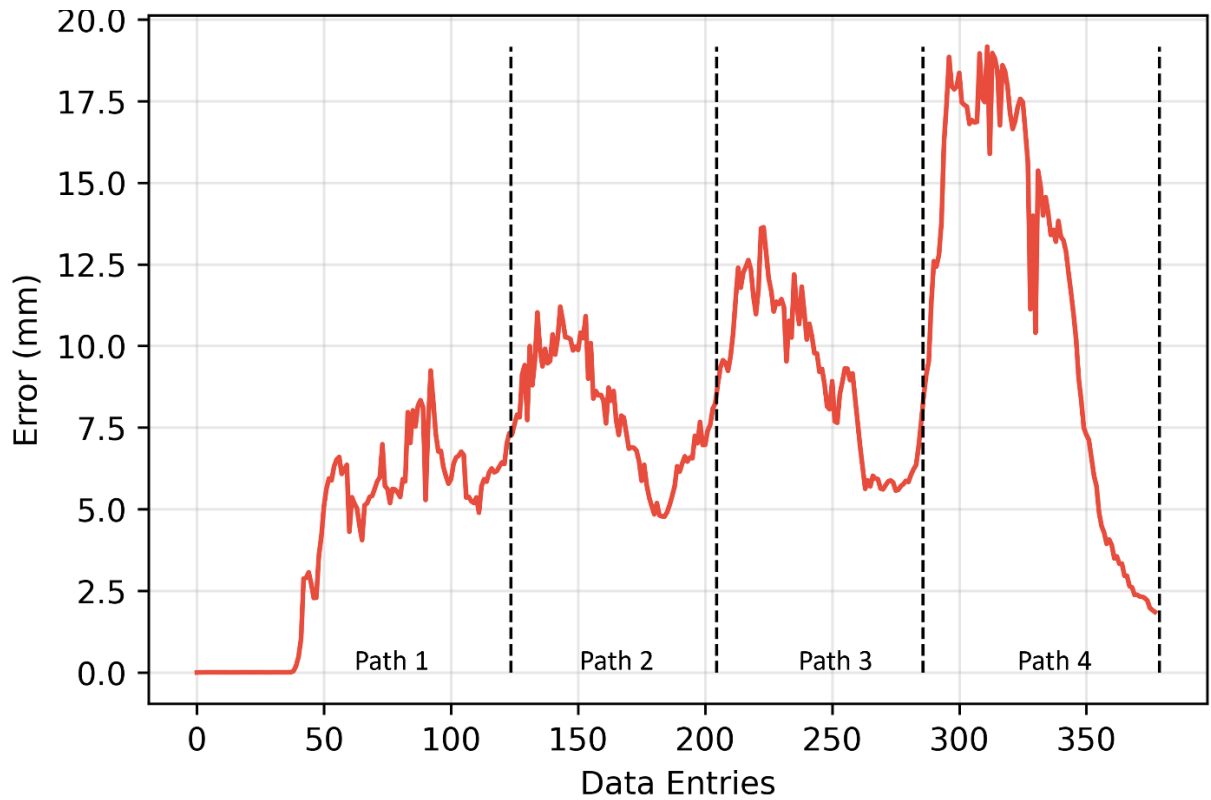

Supplementary Figure 5. Mean test error for mapped 0.16m/s.

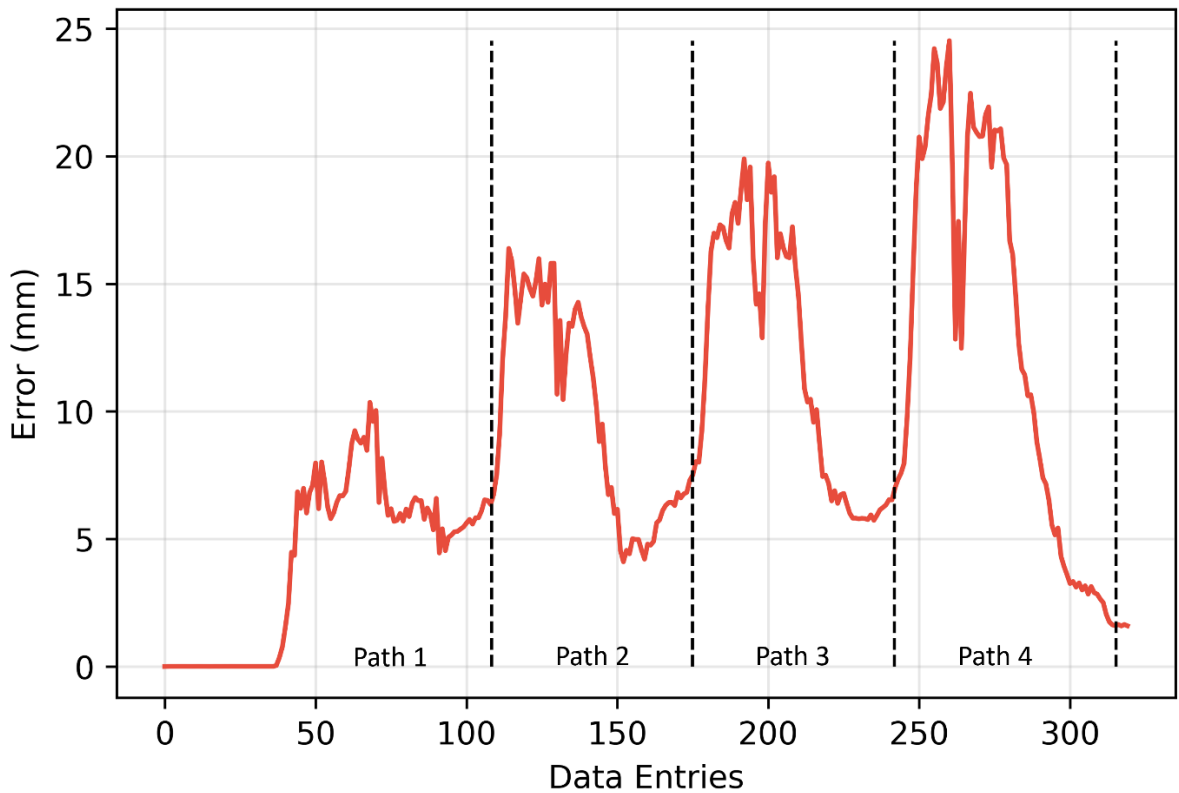

Supplementary Figure 6. Mean test error for mapped 0.24m/s.